\pdfoutput=1

\documentclass[11pt]{article}

\usepackage{acl}

\usepackage{times}
\usepackage{url}
\usepackage{amsmath}
\usepackage{times}
\usepackage{graphicx}
\usepackage{xcolor}
\usepackage{latexsym}
\usepackage[capitalize]{cleveref}
\usepackage{microtype}
\usepackage[T1]{fontenc}
\usepackage[utf8]{inputenc}
\usepackage{tabularx}
\usepackage{booktabs}
\usepackage{multirow}
\usepackage{xspace}
\usepackage{floatrow}  %
\setkeys{Gin}{width=\linewidth}

\usepackage{inconsolata}

\graphicspath{{img/}}

\makeatletter
\DeclareRobustCommand\onedot{\futurelet\@let@token\@onedot}
\def\@onedot{\ifx\@let@token.\else.\null\fi\xspace}
\def\eg{e.g\onedot} 
\def\ie{i.e\onedot} 
 
 \def\vs{vs\onedot}

\makeatother

\newcommand{\mysubsection}[1]{\vspace{0.3em}\noindent\textbf{#1}}

\title{Learning Human Action Representations from \\ Temporal Context in Lifestyle Vlogs}

\author{Oana Ignat \hspace{5pt} 
 Santiago Castro \hspace{5pt}
 Weiji Li \hspace{5pt}
Rada Mihalcea \\
University of Michigan - Ann Arbor, USA \\
\texttt{oignat@umich.edu}}

\begin{document}
\maketitle
\begin{abstract}
We address the task of human action representation and show how the approach to generating word representations based on co-occurrence can be adapted to generate human action representations by analyzing their co-occurrence in videos. To this end, we formalize the new task of human action co-occurrence identification in online videos, \ie{}, determine whether two human actions are likely to co-occur in the same interval of time.
We create and make publicly available the \textsc{Co-Act} (Action Co-occurrence) dataset, consisting of a large graph of \(\sim\)12k co-occurring pairs of visual actions and their corresponding video clips. We describe graph link prediction models that leverage visual and textual information to automatically infer if two actions are co-occurring.
We show that graphs are particularly well suited to capture relations between human actions, and the learned graph representations are effective for our task and capture novel and relevant information across different data domains.
The \textsc{Co-Act} dataset and the code introduced in this paper are publicly available at \url{https://github.com/MichiganNLP/
vlog_action_co-occurrence}.
\end{abstract}

\section{Introduction}
Action understanding is a long-standing goal in the development of intelligent systems that can meaningfully interact with humans, with recent progress made in several fields including natural language processing~\cite{Fast2016AugurMH, Wilson2017MeasuringSR, Wilson2019PredictingHA}, computer vision \cite{Carreira2017QuoVA, Shou2017CDCCN, Tran2018ACL, Chao2018RethinkingTF, Girdhar2019VideoAT, Feichtenhofer2019SlowFastNF}, data mining~\cite{Kato2018CompositionalLF, Xu2019LearningTD}, and others.  %
Many of the action understanding systems developed to date, however, rely mostly on pattern memorization and do not effectively understand the action, which makes them fragile and unable to adapt to new settings~\cite{Sigurdsson2017WhatAA, Kong2018HumanAR}.  

Effective action understanding requires reliable action representations. In this paper, we introduce a strategy to generate contextual representations for human actions by adopting an approach for creating word representations based on co-occurrence information.
In linguistics, co-occurrence is defined as an above-chance frequency of ordered occurrence of two adjacent terms in a text corpus. For example, if the concepts ``peanut butter'', ``jelly'', and ``sandwich'' appear more often together than apart, they would be grouped into a concept co-occurrence rule. Co-occurrence is a building block concept for word representations and language models.
We adapt this approach to human actions, which also have their own co-occurrence relations, expressed as temporal context. 
\begin{figure}
    \includegraphics{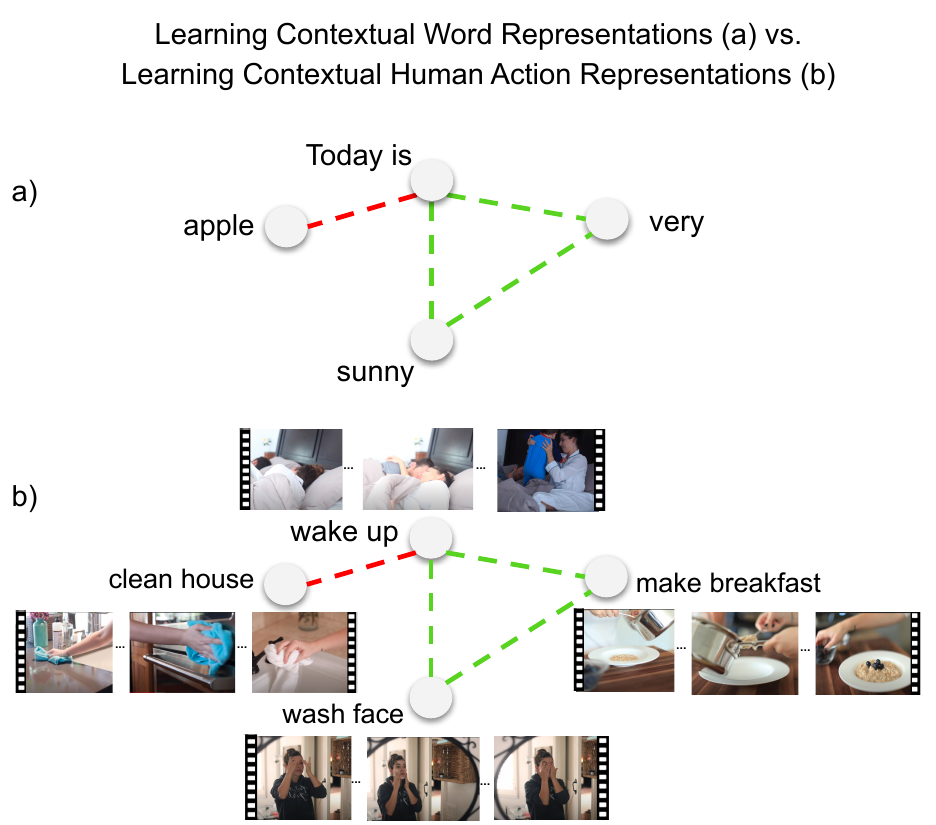}
    \caption{
    We draw inspiration from contextual word representations to create novel action representations based on video temporal context. 
    Specifically, when predicting the next word in a sentence, it is more expected to see certain words, for instance, after ``Today is'' an expected word is ``sunny'' and not ``apple''.
    Similarly, human actions also follow a certain pattern, for instance, after ``waking up'', an expected next action is to ``wash the face'' and not to ``clean the house''.}
    \label{fig:graph_intro}
    \vspace{-0.5em}
\end{figure}
Most human actions are interconnected, as an action that ends is usually followed by the start of a related action and not a random one (\eg{}, after ``waking up'', one would ``wash face'' or ``make breakfast'' and not ``sell books'' or ``go to bed''). We model this information through co-occurrence relations: in general, we expect that the actions `wake up'', ``wash face'' and ``make breakfast'' co-occur in a short interval of time, while ``wake up'', ``clean house'' or ``go to bed'' do not.
A natural way to model the connections between human actions is through a graph representation, where actions are represented as nodes, and their co-occurrences are represented as edges  (\cref{fig:graph_intro}).

The interconnection of human actions is well captured in lifestyle vlogs, where vloggers visually record their everyday routine consisting of the activities they perform during a regular day~\cite{Fouhey2018FromLV, Ignat2019IdentifyingVA, Ignat2021WhyActIA}.
We collect a dataset of lifestyle vlogs from YouTube that are currently very challenging for systems to solve.

\textbf{Contributions.} Our paper makes four main contributions. 
First, we show how the approach to generating word representations based on co-occurrence can be adapted to generating representations for human actions by analyzing their co-occurrence in videos. To this end, \textbf{we formalize the new human action co-occurrence identification task} in online videos.
Second, \textbf{we introduce a new dataset}, \textsc{Co-Act}, consisting of a large graph of co-occurring actions in online vlogs.
Third, \textbf{we propose several models to solve the task of human action co-occurrence}, by using textual, visual, multi-modal, and graph-based action representations. 
Finally, \textbf{we show that our action representations based on co-occurrence capture novel and relevant information across different data domains}, which leads to rich avenues for future work for improving action representation and making progress toward the broader goal of action understanding.

\section{Related Work}
\vspace{-2mm}

There are three areas of research related to our work: human action co-occurrence, graph link prediction and webly-supervised learning 

\paragraph{Human Action Co-occurrence.}
Recent work shows that action co-occurrence priors~\cite{Kim2020DetectingHI, Kim2021ACPAC} increase the performance of human-object interaction models and lead to more effective training, especially in long-tail classes. %
Unlike our work, they assume that the action co-occurrence information is provided and do not attempt to learn it. To the best of our knowledge, we are the first to propose the task of learning human action co-occurrence in videos. 

Human action co-occurrence identification is also related to learning action temporal order in videos which is used to construct the co-occurring action pairs.
\citet{Misra2016ShuffleAL} propose the task of temporal order verification, \ie{}, to determine whether a sequence of frames from a video is in the correct temporal order. Using this simple task and no semantic labels, they learn visual representation. 
In our work, we learn action representations using the information extracted from the action co-occurrence graph, a more general relation reflecting a shared context among the actions.%
\vspace{-0.7em}
\paragraph{Link Prediction.}
Link prediction is a key problem for graph-structured data and is relevant for our graph formulation of action co-occurrence. The objective of link prediction is to predict whether two nodes in a graph are likely to be linked~\cite{LibenNowell2007TheLP}.

Link prediction approaches can be categorized into three main categories~\cite{Kumar2020LinkPT}: similarity-based/heuristic~\cite{Newman2001ClusteringAP,JaccardEtudeCD,Salton1983IntroductionTM,Adamic2003FriendsAN,Ravasz2002HierarchicalOO,Zhou2009PredictingML,LibenNowell2007TheLP}; probabilistic-based~\cite{Kashima2006APP}; and dimensionality reduction-based (\eg{}, embedding-based or other learning approaches; \citealp{Grover2016node2vecSF,Kipf2017SemiSupervisedCW}).

For our task, we apply the similarity-based, embedding-based, and learning-based models.
Similarity-based methods are the simplest and measure similarity between every pair of nodes using topology properties of the graph (\eg{}, common neighbors). 
The embedding-based link prediction models map the embedding of nodes to a lower dimension such that similar nodes have similar embeddings.
The learning-based link prediction models can be cast using supervised classification models where a point corresponds to a node pair in the graph, and the point label represents the presence or absence of an edge/link between the pair. %

\vspace{-0.7em}
\paragraph{Webly-Supervised Learning.}
In our work, we identify human action co-occurrence in the context of rich, virtually unlimited, constantly evolving online videos from YouTube, using the video transcripts as a web supervision signal.
Large-scale video datasets on instructional videos \cite{Miech2019HowTo100MLA} and lifestyle vlogs \cite{Fouhey2018FromLV,Ignat2019IdentifyingVA, Ignat2021WhyActIA,Ignat2022WhenDI} are other examples of web supervision. The latter is similar to our work as they analyze online vlogs, but unlike ours, their focus is on action detection or the reasons behind actions and not on action co-occurrence.

\vspace{-2mm}
\section{Dataset}
\vspace{-2mm}

To develop and test models for determining if two actions co-occur, we compile a novel dataset,
which we refer to as \textsc{Co-Act} (Action Co-occurrencE).

\subsection{Data Collection}
We start by compiling a set of lifestyle videos
from YouTube, consisting of people performing their daily routine activities, such as cleaning, cooking, studying, relaxing, etc.
We build a data-gathering pipeline to automatically extract and filter videos and their transcripts. 

We select 20 YouTube channels and download all the videos and their transcripts. The channels are selected to have good-quality videos with automatically generated transcripts containing detailed verbal descriptions of the actions depicted. 

An analysis of the videos indicates that both the textual and visual information are rich sources for describing not only the actions but also in what order the actions are performed, making them a great source of data for developing action co-occurrence models. The routine nature of the videos means that
the vloggers record and describe their actions in the order they normally occur in a day: \eg{}, ``wake up'', ``make bed'', ``wash face'', ``make breakfast'', ``drive to work'', and so on. They can also choose to focus on certain activities (\eg{}, often cooking) and enumerate more fine-grained actions related to those activities (\eg{}, ``cut apple'', ``add peanut butter''). Therefore, our dataset contains both general and fine-grained actions.
We present data analyses in \cref{data_analysis}. 
\vspace{-0.6em}
\paragraph{Action extraction.}
Having a comprehensive list of actions is necessary for creating graphs that contain most of the actions in the videos. At the same time, not all the actions from the transcripts are useful, as many of them are not visible in the video or hard to detect by computer vision systems (\eg{}, ``feel'', ``talk'', ``thank'', ``hope'', ``need'', ``see').

Therefore, we first ensure that the actions we collect are mostly visible in the videos. Our strategy is to extract all the verbs from the transcripts and then filter them using a list of ``visual verbs'' collected from imSitu~\cite{yatskar2016}, COCO-a~\cite{ronchi2015describing}
and Levin~\cite{Levin1993EnglishVC}.\footnote{Levin's taxonomy provides a classification of 3,024 verbs (4,186 senses) into 48 broad and 192 fine-grained classes. We leave analyzing the Levin verb taxonomy impact on human action model performance as a future work direction.}
Verbs from imSitu and COCO-a are considered visual as the dataset collection pipelines include an explicit annotation step to determine if verbs are visual. We manually filter and check the verbs collected from Levin.

Next, we extract all actions from the video transcripts using the dependency parser from spaCy~\cite{spacy} by extracting all the verbs and their corresponding verb phrase direct objects, prepositions, and objects of prepositions. We find that extracting only verbs and their corresponding direct objects does not always return comprehensive actions (\eg{}, ``add teaspoon'' versus ``add a teaspoon of salt''). 
We also find that many verbs do not have informative direct objects (\eg{}, ``write it'', ``clean them''), which makes the actions harder to differentiate and visually recognize. To address this, we apply co-reference resolution on the video transcripts using spaCy~\cite{spacy} NeuralCoref\footnote{\url{https://spacy.io/universe/project/neuralcoref}} model and re-extract the actions from the processed transcripts.

Finally, we obtain our visible actions by filtering all the transcript-extracted actions that contain visual verbs.
\vspace{-0.6em}
\paragraph{Video extraction.}
As transcripts are temporally aligned with videos, we can obtain meaningful video clips related to the narration. We extract clips corresponding to the visual actions based on transcript timestamps. From 2,571 videos, we obtain 19,685 unique video clips and 25,057 (action, video-clip) pairs.
Note that an action can be present in multiple video clips, and conversely, a video clip can contain multiple actions.
To control the number of clips per action, we randomly sample up to 10 random video clips for each action and finally obtain 12,994 (action, video-clip) sampled pairs.
\vspace{-0.6em}
\paragraph{Quality Assurance.}
As described above, we perform multiple steps to ensure the actions appear in the videos. 
First, we manually select 20 YouTube channels from vloggers with high-quality filming styles, who usually provide detailed visual and textual descriptions of their actions.
Second, we automatically extract actions that contain visual verbs. We manually check around 100 extracted actions to see if they are parsed well and if they correctly match their corresponding video and transcript context.
Third, we automatically filter out videos that do not contain any transcripts or no significant motion.
We filter out the motionless videos by following the procedure from~\citet{Ignat2019IdentifyingVA}: we sample one out of every one hundred frames of the videos and compute the 2D correlation coefficient between these sampled frames. We filter out all the videos with a median of the values greater than a threshold (0.8).
We manually check around 100 (action, video) pairs to see if they correctly match and find around 18 complete misalignments.
Finally, to mediate the misalignment and obtain diverse filming perspectives, we randomly sample up to 10 video clips for each action, which increases the chances that the action is present in at least one video. Random examples of actions and their video-frames are found in \href{https://anonymous.4open.science/r/sample_frames}{sample-frames}.

\subsection{Data Pre-processing}%
\label{sec:data-preprocessing}

After collecting videos, transcripts, and actions, the following data pre-processing steps are applied.

\begin{table}
    \small
    \begin{tabular}{lccc}
    \toprule
     & \#Verbs & \#Actions & \#Action pairs \\
    \midrule
    Initial & 608 & 20,718 & - \\
    Co-occurrence & 439 & 18,939 & 80,776 \\
    Clustering & 172 & 2,513 & 48,934 \\
    Graph & 164 & 2,262 & 11,711 \\
    \bottomrule 
    \end{tabular}
    \caption{Statistics for the collected number of unique verbs, actions, and co-occurring action pairs at each stage of data pre-processing.}
    \label{tab:filter_stats}
\end{table}

\vspace{-0.6em}
\paragraph{Action Co-occurrence Selection.}
From all the extracted visual actions, we automatically select all the action pairs that are co-occurring.
We define two actions as co-occurring if they are less than 10 seconds away from each other. The 10 seconds is an \textit{intermediate value threshold} we set after experimenting with other values. This threshold controls the scale of time we choose to focus on when collecting co-occurring actions: \eg{}, mostly short actions (\eg{}, ``open fridge'', ``get milk'') are captured in a small interval of time (1-5 sec), while longer intervals allow for longer and more diverse actions to co-occur (\eg{}, ``prepare meal''). We choose an intermediate value that allows for both shorter and longer actions to co-occur\footnote{The captured actions also depend on the filming style (\eg{}, vloggers could increase the filming time of normally short actions).}. 
Our intuition is that modeling the relations between both shorter and longer actions would result in learning more comprehensive information about human actions. We also consider the in-depth analysis of this threshold and its downstream effects as an interesting future work direction and our framework allows for effortless threshold tune.

For computing the distance in time between two actions, we use the transcript time stamps. This allows scaling data with no constraints from the annotation budget. The transcript time stamps do not always match the time the action appears in the video. However, this hardly impacts our task because the actions mentioned in the transcript usually follow the order from the video. Furthermore, we mediate misalignments by collecting multiple videos per action and filtering steps described in the previous section.
\vspace{-0.6em}
\paragraph{Action Clustering.}
We find that many actions are often very similar in meaning. This leads to many action repetitions: \eg{}, ``use iron'', ``iron shirt'', ``iron cloth''. To avoid such repetitions, we group similar actions by clustering all actions.
We represent each action using the pre-trained model Sentence-BERT \cite{reimers-2019-sentence-bert} and apply Agglomerative Clustering~\cite{Murtagh2014WardsHA}.  We filter out the clusters of actions with less than two actions, as they are likely to be outliers that were not well extracted. The actions in each cluster are then renamed to the most common action in the cluster: \eg{}, ``iron shirt'' and ``iron cloth'' are renamed to ``use iron''.

We observe that the clustering model is introducing some noise level as it does not perfectly cluster all actions. We tried to mitigate this with different Sentence-BERT pre-trained models for sentence similarity\footnote{\href{https://sbert.net/docs/pretrained_models.html}{sbert.net/docs/pretrained\_models.html}}
and fine-tuning our clustering model hyper-parameters\footnote{linkage distance threshold (1.5), linkage criterion (ward)} based on automatic evaluation metrics for measuring the quality of clusters\footnote{Silhouette Coefficient~\cite{Rousseeuw1987SilhouettesAG}, Calinski-Harabasz Score~\cite{Caliski1974ADM}, and Davies-Bouldin Index~\cite{Davies1979ACS}}.

\vspace{-0.6em}
\paragraph{Action Graph Filtering.}
After we rename the actions based on clustering, we create a graph where the nodes represent the actions, and the edges represent the relations between two actions.
Specifically, we create an undirected
graph for each video, where the graph nodes are represented by the actions in the video and the co-occurring actions are connected by an edge. Each edge has a weight equal to the number of times the corresponding actions co-occur in the video.

We combine all the video graphs to obtain a single large graph that contains all the co-occurring actions in our data. We filter out the action pairs that co-occur only once in the graph (their edge weight equals one), as their co-occurrence relation is not strong and might be random.
We show the statistics before and after all the action filtering steps in \cref{tab:filter_stats}. 
More information (\eg{}, action frequency distributions, action pairs) can be found in \cref{appendix}.

\vspace{-1mm}
\subsection{ACE vs\onedot{} current Human Action Datasets}

Many publicly available visual action datasets ~\cite{Carreira2017QuoVA, Soomro2012UCF101AD, Kuehne11} do not have video transcripts and do not have videos with multiple actions presented in their natural order, therefore we cannot leverage the textual information and the relations between actions, as we can do in our dataset. 

The majority of human actions datasets with transcripts are restricted to one or few domains (\eg{}, cooking~\cite{Zhou2018TowardsAL} or instructional videos~\cite{Miech2019HowTo100MLA,coin}).
The main difference between lifestyle vlogs and instructional videos is the domain of the actions. Instructional videos are usually from just one domain (\eg{}, either cooking, repairing, or construction) and tend to have a specialized vocabulary (\eg{}, car repair). Lifestyle vlogs contain various everyday actions from multiple domains in the same video (cleaning, cooking, DIY, entertainment, personal care). Due to the diversity of domains in our data, our model learns not only the co-occurrence between in-domain actions (\eg{}, cooking: ``cut potato'' \& ``add onion'') but also the relations from different domains (\eg{}, personal care and cooking: ``wash face'' \& ``make breakfast'').

\vspace{-2mm}
\section{Action Co-occurrence in Vlogs}
\vspace{-2mm}

We formulate our action co-occurrence identification task as a link prediction task.
Link prediction aims to predict the existence of a link between two nodes in a graph.
In our setup, nodes are represented by actions, and every two co-occurring actions are connected by a weighted edge, where the weight represents the number of times the two actions co-occur. Our goal is to determine if an edge exists between two given actions.\footnote{At this point, we do not aim to also identify the strength of the link.} 

\subsection{Data Representation}%
\label{sec:data_proc}

\paragraph{Textual Representations.}
To represent the textual data -- actions and their transcript context, we use Sentence Embeddings computed using the pre-trained model Sentence-BERT embeddings~\cite{reimers-2019-sentence-bert} calculated using the graph topology and the textual embeddings obtained from CLIP~\cite{Radford2021LearningTV}.
When computing CLIP textual action embeddings, we concatenate the action with given prompts (\eg{}, ``This is a photo of a person''), as described in the original paper~\cite{Radford2021LearningTV}.
\vspace{-0.6em}
\paragraph{Video Representations.}
We use the CLIP model~\cite{Radford2021LearningTV} to represent all the actions and their corresponding video clips.
One action can have multiple video clips: an action has at most 10 corresponding videos.
From each video clip, we extract four equally spaced frames and pre-process them as done before ~\cite{Radford2021LearningTV}.
We use the pre-trained Vision Transformer model ViT-B/16~\cite{dosovitskiy2020vit} to encode the video frames and the textual information.
We apply the model to each of the four frames and average their representations~\cite{Luo2021CLIP4ClipAE}.
\vspace{-0.6em}
\paragraph{Graph Representations.}
We also use the training graph topology information (node neighbors and edge weights) to compute action embeddings as the weighted average of all of their neighbor node embeddings, where the weights are edge weights (\ie{}, how many times the two nodes co-occur). The neighbor node embeddings are represented using either textual embeddings (Sentence-BERT; \citealp{reimers-2019-sentence-bert}) or visual embeddings (CLIP; \citealp{Radford2021LearningTV}).
All the graph-based models described in the next section use graph topology information from the validation graph (see \cref{sec:evaluation-data-split}).

We use the representations described above as input to different action co-occurrence models.

\subsection{Action Co-occurrence Models}
We explore many models with various input representations.
We group the models as described in the related work link prediction section: random baseline, heuristic-based models (graph topology models), embedding-based models (cosine similarity and graph neural networks), and learning-based models (SVM models).
As described in \cref{sec:data_proc}, we run experiments with various types of data representations:
Textual: Action and Action Transcript; Visual: Action, Video, and Multi-modal (Action\&Videos; the average between action and video visual embeddings); Graph: Action and Multi-modal (Action\&Videos) using graph topology.

\subsubsection{Random Baseline}
The action pairs to be predicted as co-occurring or not are split into equal amounts, therefore a random baseline would have an accuracy score of 50\%.

\subsubsection{Heuristic-based Graph Topology Models}
We apply several popular node similarity methods that only use graph topology information in the prediction process: Common Neighbours~\cite{Newman2001ClusteringAP}, 
Salton Index~\cite{Salton1983IntroductionTM}, 
Adamic-Adar Index~\cite{Adamic2003FriendsAN}, 
Hub Promoted Index~\cite{Ravasz2002HierarchicalOO}, 
and Shortest Path~\cite{LibenNowell2007TheLP}.
Note that the heuristic-based methods do not use any data representations described in \cref{sec:data_proc}.
We describe each of the methods above:

\paragraph{Notation.} 
Let \(s_{xy}\) be the similarity between nodes \(x\) and \(y\), \(\Gamma(x)\) be the set of nodes connected to node \(x\) and \(k_{x}\) be the degree of node \(x\).

\paragraph{Common Neighbours.}
Two nodes are more likely to be connected if they have more common neighbors.
\begin{equation}\small 
    s_{xy} = |\Gamma(x) \cap \Gamma(y)|
\end{equation}

\paragraph{Salton Index.}
Measures the cosine of the angle between columns of the adjacency matrix corresponding to given nodes. 
\begin{equation}\small 
    s_{xy} = \frac{|\Gamma(x) \cap \Gamma(y)|}{\sqrt{k_x k_y}}
\end{equation}

\paragraph{Hub Promoted Index.}
This measure assigns higher scores to edges adjacent to hubs (high-degree nodes), as the denominator depends on the minimum degree of the nodes of interest.
\begin{equation}\small 
    s_{xy} = \frac{|\Gamma(x) \cap \Gamma(y)|}{\min\{k_x, k_y\}}
\end{equation}

\paragraph{Adamic-Adar Index.}
This measure counts common neighbors by assigning weights to nodes inversely proportional to their degrees. That means that a common neighbor, which is unique for a few nodes only, is more important than a hub.
\begin{equation}\small 
    s_{xy} = \sum_{z \in \Gamma(x) \cap \Gamma(y)} \frac{1}{\log{k_z}}
\end{equation}

\paragraph{Shortest Path.}
The similarity score is inversely proportional to the length of the shortest path between two nodes. 
\begin{equation}\small 
    s_{xy} = \frac{1}{min\{l : path_{xy}^{<l>} exists\}}
\end{equation}

\paragraph{Weighted Graph Models.}
Our graph is weighted, therefore we also apply weighted graph models. We modify some of the above models (Common Neighbours, Adamic-Adar Index) to use the link weight information, as proposed in~\citet{Zhu2016LinkPI}. We find that using link weights achieves similar results as without them.

\subsubsection{Embedding-based Models}
\paragraph{Cosine Similarity.}
We compute the cosine similarity between their embeddings to determine if two given actions co-occur. If the similarity score is greater than a threshold fine-tuned on validation data, we predict the actions as co-occurring.
\vspace{-0.6em}
\paragraph{Graph Neural Networks.}
We also use Graph Neural Network (GNN) models. We choose four diverse and popular models~\cite{Kumar2020LinkPT}: 
attri2vec~\cite{Zhang2019AttributedNE}, GraphSAGE~\cite{Hamilton2017InductiveRL},  GCN~\cite{Kipf2017SemiSupervisedCW}.
GNN models can also be classified as learning-based models: they learn a new heuristic from a given network, as opposed to Graph Topology models, which use predefined heuristics, \ie{}, score functions. 
We create our graph based on a known heuristic: co-occurring actions are closely connected in the graph. Therefore, we hypothesize that heuristic models will perform better.
Indeed, we observe that for our graph, the GNN methods do not perform better than the heuristic models: the best-performing model is GraphSAGE with 77.2\% accuracy, while the best-performing topology model has an 82.9\% accuracy (see \cref{tab:eval}). Therefore, we conclude that our task does not benefit from these neural models.

\subsubsection{Learning-based Model}
We run a support vector machine (SVM)~\cite{cortes1995support} classifier on each action pair to be classified as co-occurring.
We concatenate all the input representations and the heuristic scores, and we standardize the features by removing the mean and scaling to unit variance. We fine-tune the model hyper-parameters (kernel, C, gamma) on the validation data using a grid search.

\vspace{-2mm}
\section{Evaluation}
\vspace{-2mm}

We conduct extensive experiments to evaluate the action pairs co-occurrence identification task. The task can be represented as a graph link prediction task. 
Therefore, we adopt the link prediction evaluation process. 

\subsection{Evaluation Data Split}%
\label{sec:evaluation-data-split}
We split the original graph into train, validation, and test graphs.
In link prediction, the goal is to predict which links will appear in the future of an evolving graph. 
Therefore, while keeping the same number of nodes as the original graph, the number of edges is changed as some of the edges are removed during each split and used as the positive samples for training, fine-tuning, and testing the link prediction models. 
The edges are split into the train, validation, and test sets using a transductive split, which is considered the default evaluation splitting technique for link prediction models~\cite{Xu2018HowPA}.
Specifically, we randomly sample 10\% of all existing edges from the original graph as positive testing data and the same number of nonexistent edges (unconnected node pairs) as negative testing data. The reduced graph becomes the test graph and, together with the set of sampled edges, is used for testing the models. We repeat the same procedure to create the validation and the train data for the models. The validation graph is created by reducing the test graph, and the training graph is created by reducing the validation graph. 

\setlength{\tabcolsep}{1em} 
\begin{table}
    \small
    \begin{tabular}{c|c }
    \toprule
    Model & Accuracy \\
    \midrule
    \midrule
    \multicolumn{2}{c}{\sc Baseline} \\
    \midrule
    \midrule
    Random & 50.0 \\
    \midrule
    \midrule
    \multicolumn{2}{c}{\sc Heuristic-based} \\
    \midrule
    \midrule
    Common Neighbours & 82.9 \\
    Salton Index & 71.2 \\
    Hub Promoted Index & 78.3 \\
    Adamic-Adar Index & 82.9 \\
    Shortest Path & 82.9 \\
     \midrule
    \midrule
    \multicolumn{2}{c}{\sc Embedding-based} \\
    \midrule
    \midrule
   Cosine similarity & 82.8\\
  \midrule
  attri2vec & 65.7\\
  GCN & 77.2\\
  GraphSAGE & 78.1\\
    \midrule
    \midrule
    \multicolumn{2}{c}{\sc Learning-based} \\
    \midrule
    \midrule
     SVM & \textbf{91.1} \\
    \bottomrule
    \end{tabular}
    \caption{Accuracy results for all the models.}
    \label{tab:eval}
\end{table}

\setlength{\tabcolsep}{0.7em} 
\begin{table*}
    \small
    \begin{tabular}{c|c|c|c|c|c|c|c}
    \toprule
    \multirow{3}{*}{Model}
     & \multicolumn{7}{c}{\sc Input Representations} \\
    \cmidrule{2-8}
     & \multicolumn{2}{c|}{Textual} & \multicolumn{3}{c|}{Visual} & \multicolumn{2}{c}{Graph}\\
    \cmidrule{2-8}
    & Action & Transcript & Action & Video & Action\&Video & Action & Action\&Video \\
    \midrule
    Cosine Similarity & 60.6 & 65.2 & 62.7 & 57.0 & 65.4 & \textbf{82.8} & 50.6 \\
    SVM & 76.3 & 71.1 & 73.1 &  76.2 &  76.1 & 80.9 & 74.6\\
    \bottomrule
    \end{tabular}
    \caption{Ablations and accuracy results on test data. We compute the ablations for each input representation: textual, visual, and graph, for an embedding-based model (cosine similarity) and a learning-based model (SVM); the heuristic-based models do not depend on input representation type, therefore we do not ablate them.}
    \label{tab:eval2}
\end{table*}

\subsection{Results and Ablations}
\Cref{tab:eval} contains the results, measured by accuracy, for each model type. The learning-based model, SVM, using all input representations (textual, visual, graph) and all graph heuristic scores obtains the highest accuracy score. Therefore, using both graph topology information and textual embeddings leads to the best performance for our task.
The results for each of the heuristic-based graph-topology models are shown in \cref{tab:eval}.
Simple heuristics (common neighbors or shortest path) are enough to perform well. 

\paragraph{Modality Ablation.}
The ablation results, split by input representation are shown in \cref{tab:eval2}.
We analyze how different input representations influence the model's performance: textual (Sentence-BERT and CLIP textual) \vs{} visual (CLIP visual) \vs{} multi-modal (CLIP textual and visual) \vs{} graph (Sentence-BERT and CLIP textual and visual). The input representations are described in \cref{sec:data_proc}.
The textual embeddings are a strong signal for our task, even when not using any graph information: SVM with only Action Sentence-BERT embeddings has a 76.3\% accuracy. Using graph representations or graph heuristic information leads to significantly better performance (80.9\% and 91.1\% accuracy, respectively).
The visual and multi-modal embeddings are also valuable but perform worse than the textual embeddings. We hypothesize that CLIP embeddings might be affected by the time misalignment between the transcript and the video.
However, the visual modality offers important information about human actions and can be used in future work with more robust visual models.

\vspace{-1mm}

\subsection{Downstream Task: Action Retrieval}
Similar to how word embeddings have been used for word similarity and for retrieving similar words and documents~\cite{Mikolov2013EfficientEO, Devlin2019BERTPO}, our graph dataset enables \textit{action similarity} and \textit{similar action retrieval} leveraging action-specific properties in the multi-modal space.

\begin{figure}
    \includegraphics{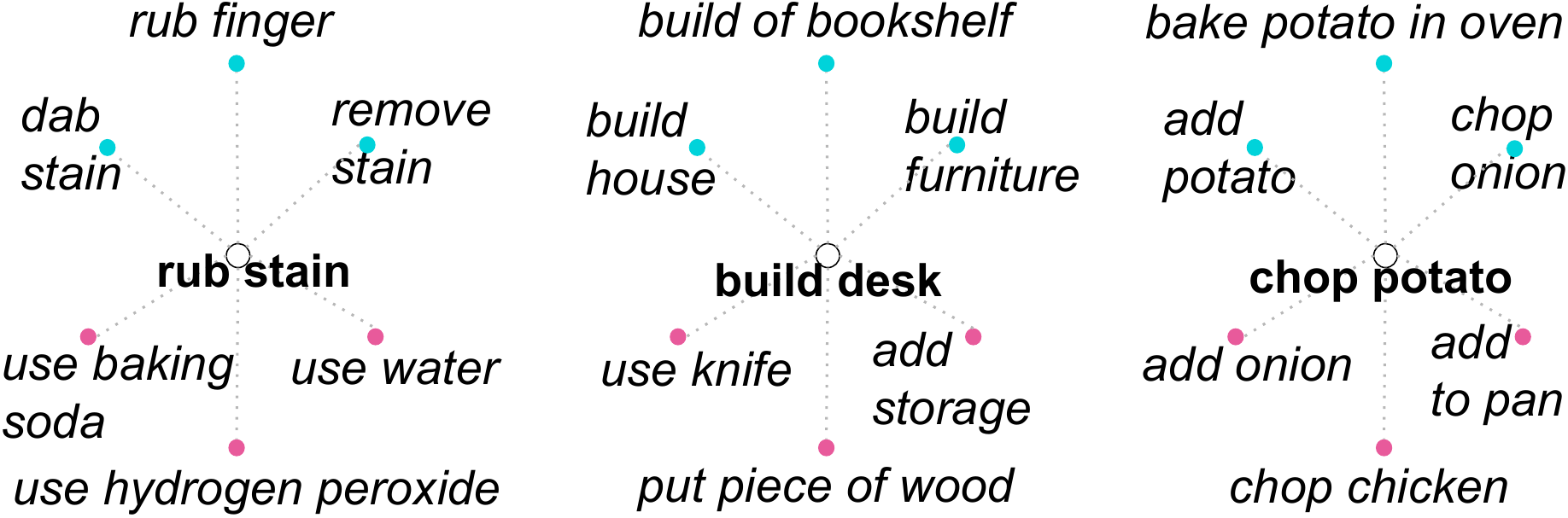}
    \caption{Top three action neighbors, obtained from textual (blue) and graph-based (purple) representations, for three random action queries from our dataset: ``rub stain'', ``build desk'', ``chop potato''.}
    \label{fig:KNN}
\end{figure}

To show the usefulness of our graph-based action embeddings, we test them on the \textit{similar action retrieval} downstream task. 
Specifically, we compare two action representations: textual (Action Sentence-BERT embeddings) and graph-based (graph weighted average of neighbor nodes Action Sentence-BERT embeddings). 
In \cref{fig:KNN}, we show the top three nearest neighbor actions from each of the representations for three random action queries from our dataset. 
We observe that \textit{each representation captures different types of information}. 
The actions obtained with textual representations are more syntactically similar to the action query, sharing either the verb or the object. This can be undesirable, as many retrieved actions are too repetitive and not always relevant to the action query:  \eg{}, ``build desk'': ``build bookshelf'', ``build house''.
In contrast, the actions obtained with graph representations are more \textit{diverse} and capture \textit{location information}, \ie{}, actions expected to be temporally close in a video: \eg{}, ``build desk'': ``use knife'', ``add storage'', ``put piece of wood''.

\paragraph{Novelty vs\onedot{} Relevance in Action Retrieval.}
A major focus in the field of Information Retrieval has been the development of retrieval models that maximize both the relevance and the novelty among higher-ranked documents~\cite{Carbonell1998TheUO}.
For the task of action retrieval, we can approximate \textit{relevance} through the \textit{location} relevance of an action, and \textit{novelty} through the \textit{diversity} of the actions retrieved.%

\paragraph{Diversity in Action Representations.}
Similar to word or document retrieval, diversity in action retrieval reflects novel results. To measure the \textit{diversity} captured by the action representations, we compute the \textit{overlap score} as the number of overlapping words between the action query and the retrieved top \(k\) action nearest neighbors, divided by the total number of words in the retrieved actions. For example, in \cref{fig:KNN}, the action query ``chop potato'', for \(k=3\), the action kNNs using textual representations (in blue) have 3 overlapping words (``chop'', ``potato'', ``potato''), from a total of 8 words, resulting in an overlap score of 3/8. We average the overlap scores across all the action queries in our dataset (2,262 unique actions) for \(k \in {3,5,10}\).
\Cref{tab:knn} shows that the actions retrieved using our graph representations have around three times fewer overlapping words with the action query; \ie{}, they are more diverse than those retrieved using the textual representation.

\begin{table}\small
    \begin{tabular}{c|c|c}
    \toprule
    \multirow{3}{*}{\(k\)} & \multicolumn{2}{c}{\sc Input Representations}\\
    \cmidrule{2-3}
    & Textual & Graph \\
    \cmidrule{2-3}
    \cmidrule{2-3}
    & \multicolumn{2}{c}{\sc Diversity/ Overlap Score \(\downarrow\)} \\
    \midrule
    \midrule
     3 & 0.35 & \textbf{0.12} \\
     5 & 0.31 & \textbf{0.11} \\
     10 & 0.26 & \textbf{0.10}\\
     \midrule
    \midrule
    Dataset &
    \multicolumn{2}{c}{\sc Location / Recall Score \(\uparrow\)}\\
    \midrule
    \midrule
    Breakfast & 0.16 & \textbf{0.22}\\
    COIN & 0.23 & \textbf{0.60}\\
    EPIC-KITCHENS & 0.14 & \textbf{0.26}\\
    \bottomrule
    \end{tabular}
    \caption{Scores measuring the difference of information, diversity, and location, between the action kNNs using different types of embeddings: textual and graph-based.}
    \label{tab:knn}
\end{table}

\vspace{-0.6em}
\paragraph{Location in Action Representations.}
To quantify how much \textit{location} information an action representation holds, we use three annotated action localization datasets%
: COIN~\cite{coin}, which includes instructional videos; EPIC-KITCHENS~\cite{Damen2018EPICKITCHENS}; and Breakfast~\cite{Kuehne12, Kuehne16end}.

We use the training data to create an action co-occurrence graph and learn action graph representations and the testing data to test our action representations.
For each action query in the test set, we obtain the actions localized before and after as the gold standard action neighbors. We also calculate the predicted action kNNs (\(k=3\)) of the action query using textual and graph-based representations. 
To measure the location information, we compute the recall score between the gold standard action temporal neighbors and the predicted action kNNs.
\Cref{tab:knn} shows that graph-based representations hold more location information than textual representations. 
Action representations that capture location information would likely benefit models in many computer vision applications, such as action localization, segmentation, or detection.\footnote{For more on how our graph can be used in other downstream tasks, see \cref{downstream}.} This leads to future research directions 
for effectively utilizing graph-based representations and co-occurring actions.

\subsection{Data Analysis}\label{data_analysis}
We want to determine which actions co-occur the most in our dataset, as it may be valuable knowledge for action recognition and action prediction systems.
Systems enriched with this knowledge can make more informed decisions when predicting or recognizing actions.
Specifically, action recognition systems can discard actions that are unlikely to happen given a previous action and assign a higher probability to actions known to co-occur with the previous action (\eg{}, given a previously recognized action ``wake up'', a likely next action could be ``wash face'', and not ``clean house'').

Given two actions, we compute their co-occurrence score using the Positive Pointwise Mutual Information (PPMI)~\cite{Church1989WordAN}. 
PMI is biased towards infrequent words, therefore we do not compute PMI for infrequent actions (that appear less than 10 times).
\begin{equation}\small 
    PPMI_{a_i,a_j} = \max(\log{\frac{P_{a_i,a_j}}{P_{a_i} P_{a_j}}}, 0)
\end{equation}
\begin{equation}\small 
    P_{a_i,a_j} = \frac{\#(a_i, a_j)}{\#action\ pairs}, P_{a_k} = \frac{\#a_k}{\#actions}
\end{equation}

\begin{figure}
    \includegraphics{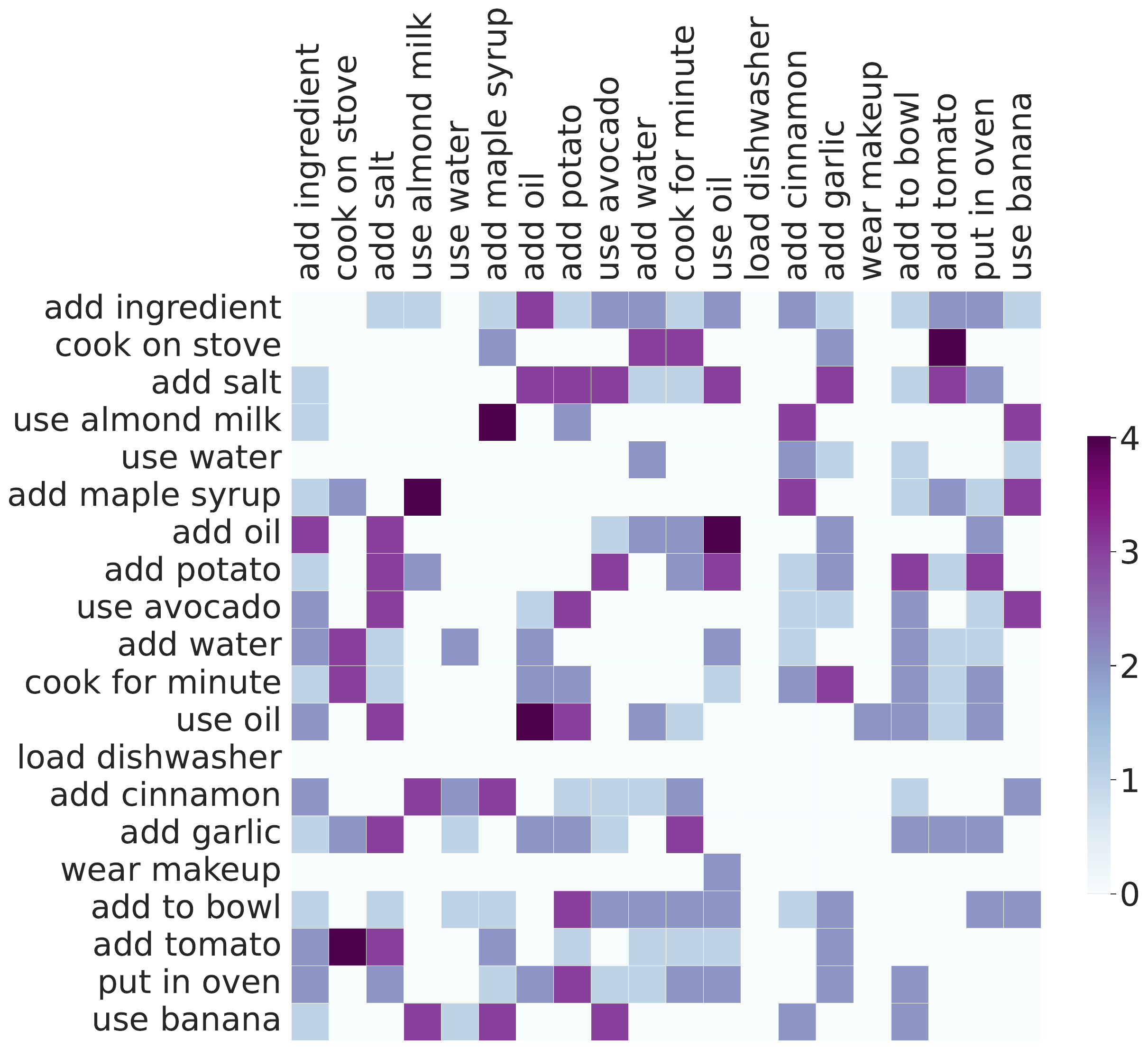}
    \caption{Co-occurrence matrix for the top 20 most frequent actions in our dataset, \textsc{Co-Act}. The scores are computed using the PPMI measure: actions with higher scores have a stronger co-occurrence relation and vice-versa. For better visualization, we sort the matrix rows to highlight clusters. Best viewed in color.}
    \label{fig:co-occurrence20}
\end{figure}

\vspace{5pt}
\Cref{fig:co-occurrence20} shows the co-occurrence matrix for the top 20 most frequent actions. The most frequent actions are related to cooking. We can see how actions related to adding ingredients are co-occurring among themselves (\eg{}, ``add potato'' and ``add avocado'') or with actions related to adding something to a container (\eg{}, ``add potato'' and ``add to bowl'').
\Cref{appendix} includes additional information: co-occurrence matrices of the top 50 most frequent actions and verbs (\cref{fig:co-occurrence_actions_50,fig:co-occurrence_verbs_50}),
top 20 actions and verb pairs co-occurring the most/least (\cref{tab:frequency-co-occurrence}),
actions and verbs distributions (\cref{fig:action-distrib,fig:verb-distrib}), top 10 most frequent clusters (\cref{fig:cluster}).

\section{Conclusion}

In this paper, we addressed the task of learning human action representations from co-occurring human actions in videos. We explored the genre of lifestyle vlogs and constructed \textsc{Co-Act}, a new dataset of \(\sim\)12k pairs of visual actions and their corresponding clips. We evaluated models that leverage textual, visual, multi-modal, and graph information.
We built \textsc{Co-Act} and action co-occurrence identification models to capture human action relations, which leads to progress towards the goal of action understanding. We are the first to address this problem and to use graph representations in this setting. We showed that graph representations are useful for our task, capture information about human actions across diverse domains, and complement the representations learned from current language and visual models.
The \textsc{Co-Act} dataset and code are available at {\url{{https://github.com/MichiganNLP/
vlog_action_co-occurrence}}.

\section*{Ethics and Broad Impact Statement}
Our dataset contains public YouTube vlogs, in which vloggers choose to share episodes of their daily life routine.
We use the videos to detect co-occurring actions without relying on information about the person's identity, such as gender, age, or location. 

The data can be used to better understand people's lives by looking at their daily routines and in which order they choose to perform their actions. The data contains videos of men and women and sometimes children, but most videos come from women. 
The routine videos present mostly ideal routines and are not comprehensive about all people's daily lives. Most of the people represented in the videos are middle-class Americans, and the language spoken is English.

In our data release, we only provide the YouTube URLs of the videos, so the creator of the videos can always have the option to remove them. YouTube videos are a frequent source of data in research papers \cite{Miech2019HowTo100MLA,Fouhey2018FromLV,AbuElHaija2016YouTube8MAL}, and we followed the typical process used by all this previous work of compiling the data through the official YouTube API and only sharing the URLs of the videos. We have the right to use our dataset how we use it, and we bear responsibility in case of a violation of rights or terms of service.

\section*{Limitations}

\mysubsection{Weak supervision from video transcripts.}
We use the weakly supervised time signal from automatically generated video transcripts without manual annotations.
This allows for no limits in scale at the cost of some noise. To reduce the noise, we use multiple (up to 10) videos to obtain the temporal action information and perform various filtering steps described in the Quality Assurance subsection. 
Furthermore, the time information is used only to find the co-occurrence information between actions, not the actual time location of the actions; therefore, it is not necessary to be clear-cut.%

\mysubsection{Directed vs\onedot{} Undirected graph representations.}
A directed graph also captures the order between the actions, which can be used in a future work direction for action prediction applications. However, an undirected graph is sufficient to obtain co-occurrence information, which suits our goal for our paper.
We looked into transforming our graph into a directed one. However, we could not do this reliably because the transcripts do not preserve the exact order of the actions. This is due to how vloggers choose to verbally describe their routines: \eg{} from ``during washing my face, I will wake up'' - it is not trivial to automatically extract the correct/natural order of the actions, as in this case, the result would be incorrect (wash face, then wake up). We tried modeling this using time keywords (\eg{}, ``during'', ``after'', ``before'') but due to the complexity of natural language, we found exceptions and other complex scenarios that could not be modeled automatically.

\mysubsection{More advanced multimodal fusion techniques.}
More advanced multimodal fusion techniques might improve the performance, and we also include this direction in future work. However, in this paper, we focused on data collection and providing a set of comprehensive link prediction baselines: heuristic-based, embedding-based, and learning-based. These baselines are challenging, as demonstrated by the high accuracy results from \cref{tab:eval}.

\bibliography{anthology}

\appendix

\section{Appendix}\label{appendix}

\subsection{How to use the graph in other datasets and downstream tasks}\label{downstream}

\paragraph{Our comprehensive graph can be extended to new data and new tasks.}

We intentionally included a large number of actions in our graph, making it comprehensive and exhaustive (see \cref{tab:filter_stats}: 164 verbs/ 2,262 unique actions), precisely to increase the chance that actions from new data can be found in our graph. 

We do not need to create a new graph for each new data or new task. Instead, we can directly use our provided learned action representations, which can also be fine-tuned on the new data. 
If there is a sufficiently similar (i.e., based on cosine similarity) class match between the actions we provide and the actions from the new data, then we can use our corresponding learned action representations. Otherwise, the new actions can be added to the graph, and new action representations can be computed. We provide code and guidelines on how to extend our graph and obtain new action representations.

\begin{figure*}
    \includegraphics[width=\columnwidth]{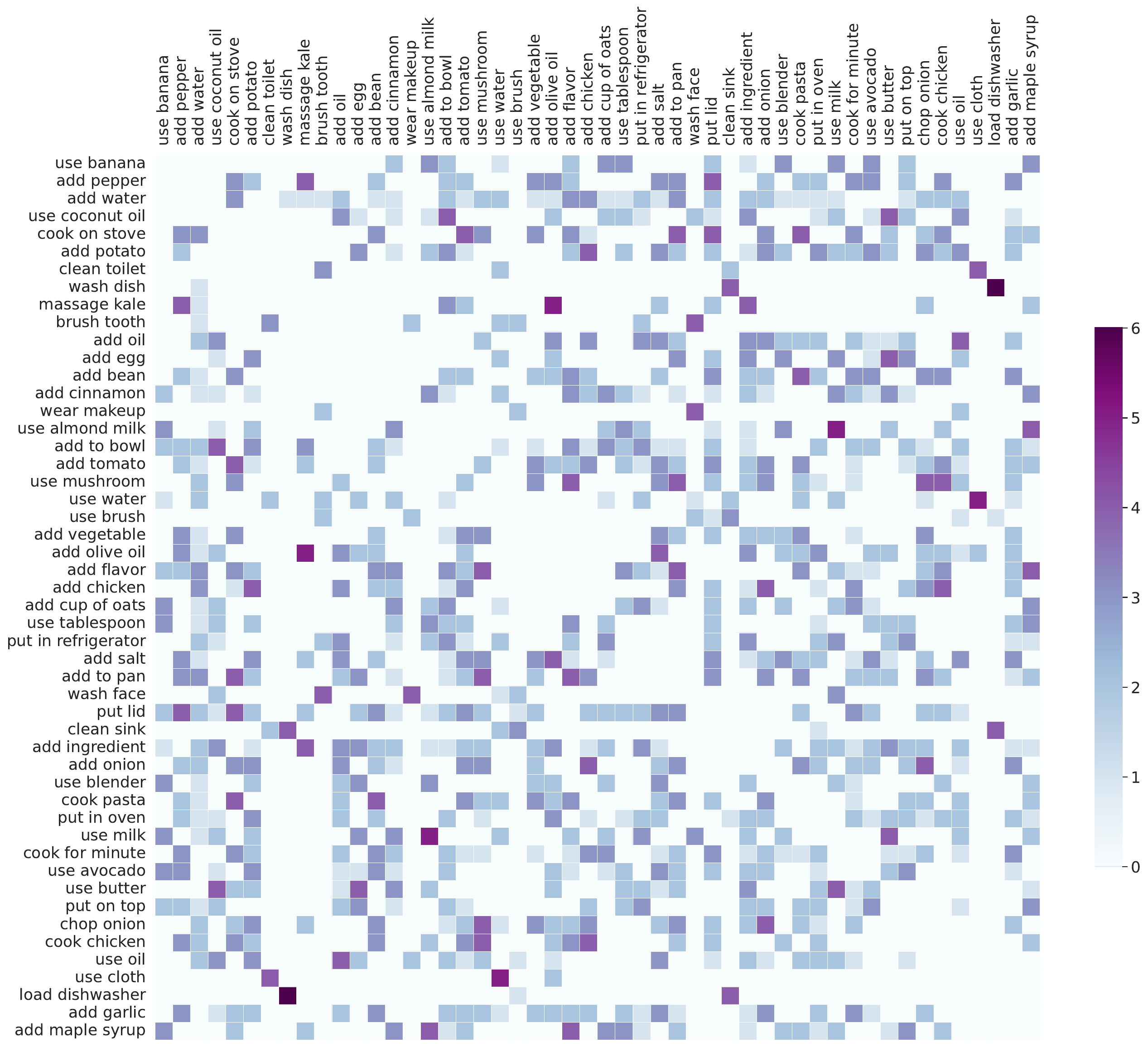}
    \caption{Co-occurrence matrix for the top 50 most frequent actions in our dataset, \textsc{Co-Act}. The scores are computed using the PPMI measure: actions with higher scores have a stronger co-occurrence relation and vice-versa. For better visualization, we sort the matrix rows to highlight clusters.
    Best viewed in color.}
    \label{fig:co-occurrence_actions_50}
\end{figure*}

\begin{figure*}
\centering
    \includegraphics[width=\columnwidth]{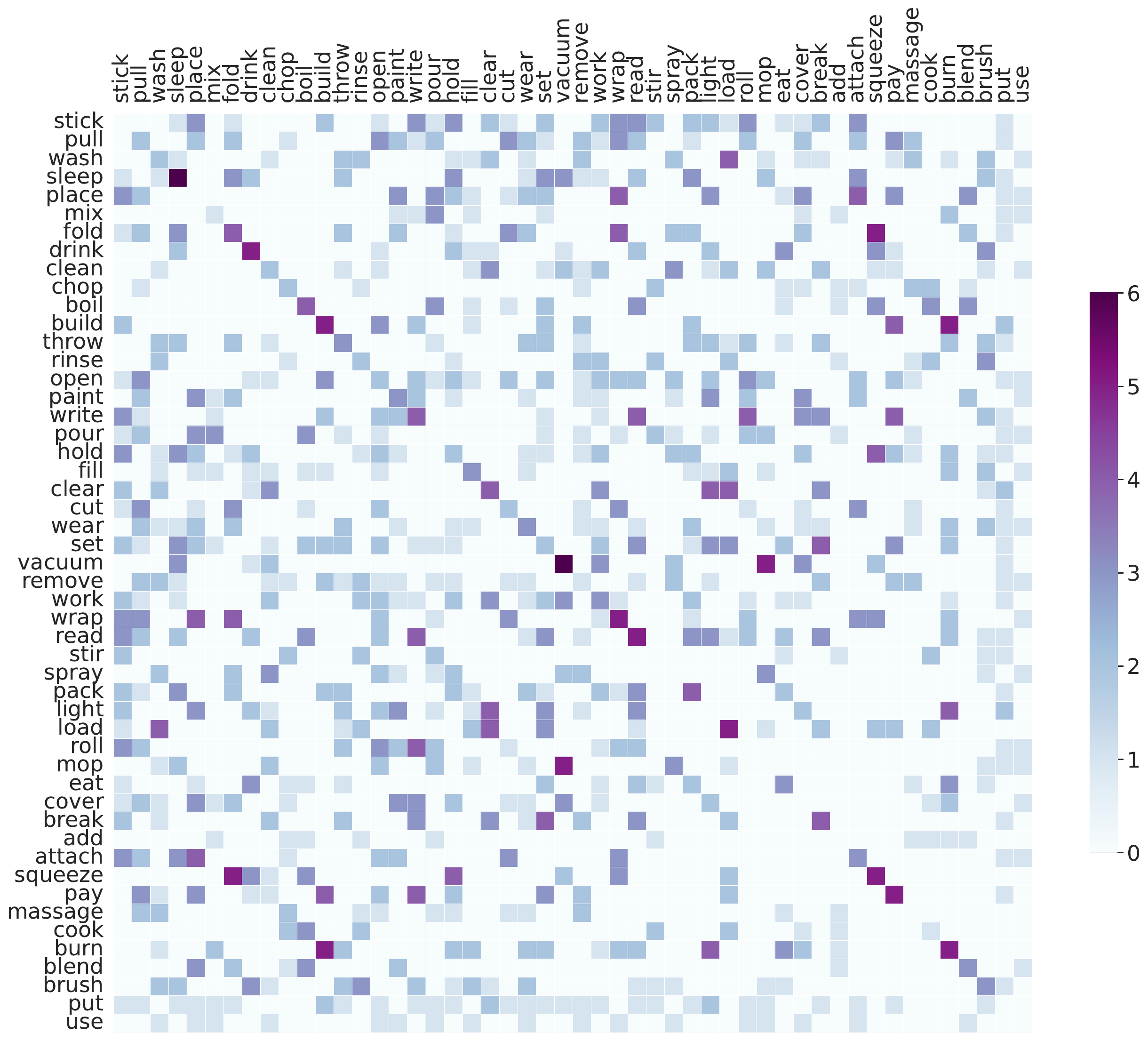}
    \caption{Co-occurrence matrix for the top 50 most frequent verbs in our dataset, \textsc{Co-Act}. The scores are computed using the PPMI measure: actions with higher scores have a stronger co-occurrence relation and vice-versa. For better visualization, we sort the matrix rows to highlight clusters.
    Best viewed in color.}
    \label{fig:co-occurrence_verbs_50}
\end{figure*}

\clearpage

\begin{minipage}[htb]{\columnwidth}%
      \centering
      \small
        \begin{tabular}{c|c}
        \toprule
        Action pair & Frequency \\
        \midrule
        load dishwasher, wash dish & 52 \\ 
        eat food, eat in day & 29 \\ 
        use shampoo, wash hair & 26 \\ 
        use cloth, use water & 24 \\ 
        add sweetener, add teaspoon of maple syrup & 23 \\ 
        use almond milk, use milk & 22 \\ 
        use butter, use purpose flour & 22 \\ 
        add olive oil, massage kale & 22 \\ 
        load dishwasher, load dishwasher at night & 22 \\
        clean steel appliance, use cloth & 21 \\ 
        put dish, wash dish & 19 \\ 
        clean toilet, spray toilet & 19 \\ 
        clean sink, use dish soap & 19 \\ 
        add cocoa powder, use purpose flour & 17 \\ 
        squeeze lemon juice, use lemon & 17 \\ 
        brush tooth, wash face & 16 \\ 
        curl eyelash, use mascara & 15 \\ 
        put in freezer, put in smoothie & 15 \\ 
        add tomato, cook on stove & 15 \\ 
        clean bathtub, use broom & 15 \\ 
         ... & ... \\
        pack makeup bag with, put in ziploc bag & 2 \\ 
        put on skin, use for lip & 2 \\ 
        put stuff, use on cuticle & 2 \\ 
        put under eye, use on cuticle & 2 \\ 
        put on eyelid, use on cuticle & 2 \\ 
        fill brow, use on cuticle & 2 \\ 
        read book, use business card & 2 \\ 
        spray paint, use iron & 2 \\ 
        use product, use vegetable peeler & 2 \\ 
        teach responsibility, work in beauty industry & 2 \\ 
        use charcoal scrub, use scrub & 2 \\ 
        use charcoal scrub, use vegetable peeler & 2 \\ 
        use charcoal scrub, use steamer & 2 \\ 
        add tea to water, use charcoal scrub & 2 \\ 
        open pore, use charcoal scrub & 2 \\ 
        use charcoal scrub, use sheep mask from store & 2 \\ 
        use on drugstore, use product & 2 \\ 
        break surface of water, remove makeup & 2 \\ 
        brush hair, spray with hairspray & 2 \\ 
        fill brow, fill browser bed & 2 \\ 
        \bottomrule
    \end{tabular}
    \end{minipage}%
    \begin{minipage}[htb]{\columnwidth}%
    \centering
    \small
    \label{tab:frequency-co-occurrence}
        \begin{tabular}{c|c}
        \toprule
        Verb pair & Frequency \\
        \midrule
         add, use & 3864 \\ 
        use, use & 2987 \\ 
        add, add & 2895 \\ 
        put, use & 1786 \\ 
        add, put & 1060 \\ 
        add, cook & 814 \\ 
        clean, use & 724 \\ 
        put, put & 620 \\ 
        use, wear & 366 \\ 
        add, chop & 355 \\ 
        clean, clean & 330 \\ 
        cut, use & 328 \\ 
        use, wash & 317 \\ 
        add, eat & 293 \\ 
        cook, use & 284 \\ 
        add, cut & 256 \\ 
        clean, put & 246 \\ 
        eat, use & 244 \\ 
        eat, eat & 201 \\ 
        fill, use & 191 \\ 
         ... & ... \\
        bake, pull & 2 \\ 
        bake, stick & 2 \\ 
        pack, pull & 2 \\ 
        empty, hold & 2 \\ 
        brush, mix & 2 \\ 
        attach, paint & 2 \\ 
        pour, wrap & 2 \\ 
        fight, wash & 2 \\ 
        drink, massage & 2 \\ 
        add, poke & 2 \\ 
        stick, stir & 2 \\ 
        fill, scrape & 2 \\ 
        carve, cover & 2 \\ 
        curl, open & 2 \\ 
        curl, rinse & 2 \\ 
        fill, pump & 2 \\ 
        build, draw & 2 \\ 
        teach, work & 2 \\ 
        break, remove & 2 \\ 
        brush, spray & 2 \\ 
        \bottomrule
    \end{tabular}
    \end{minipage}
    Top 20 most and least frequent action pairs (left) and verb pairs (right) in our dataset.

\clearpage

\subsection{Action and Verb Distribution}

\begin{figure}
    \includegraphics{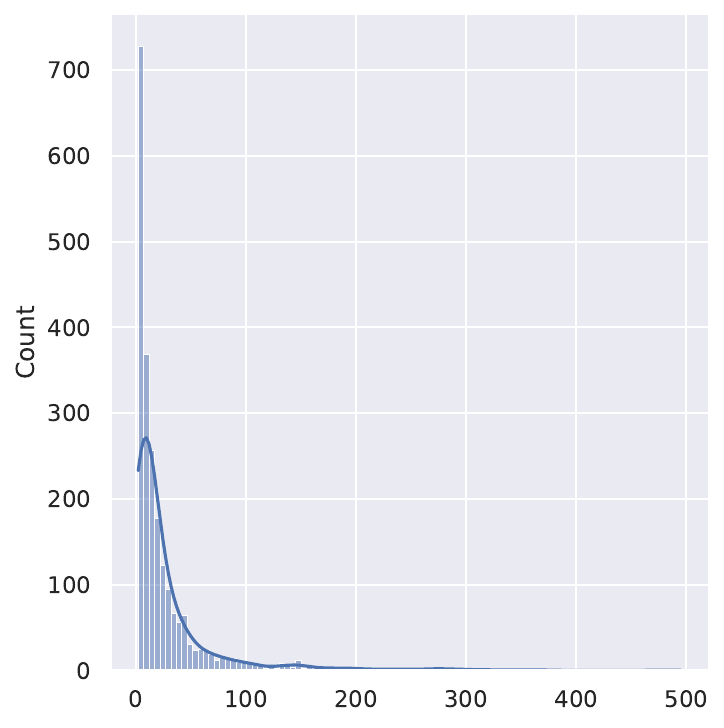}
    \caption{Action distribution in our dataset, \textsc{Co-Act}: count of actions frequencies.}
    \label{fig:action-distrib}
\end{figure}

\begin{figure}
    \includegraphics{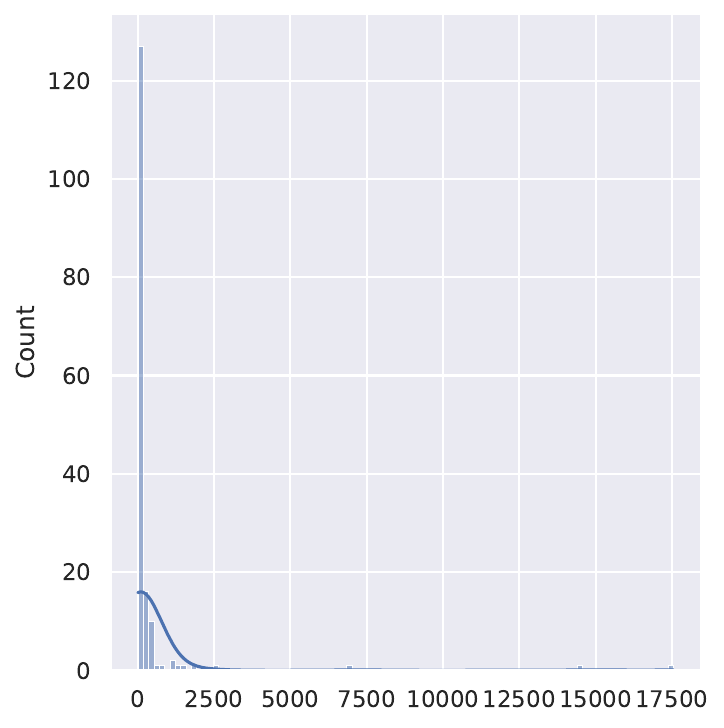}
    \caption{Verb distribution in our dataset, \textsc{Co-Act}: count of verb frequencies.}
    \label{fig:verb-distrib}
\end{figure}

\subsection{Action Clustering}
Recall that all the raw actions extracted from the transcript are clustered as described in \cref{sec:data-preprocessing}. 
To analyze the content of the clusters, we show the 10 most frequent clusters
using t-distributed Stochastic Neighbor Embedding (t-SNE)~\cite{Maaten08visualizingdata} (see~\cref{fig:cluster})

By examining the clusters, we can distinguish some open challenges or future work directions.
First, there are multiple ways of expressing the same action, which can be seen when looking at the actions inside each cluster (\eg{}, ``add to bowl', ``add into bowl'', ``place in bowl'', ``use measuring bowl''). This showcases the complexity of language.
Second, the cluster algorithms are not perfect and some clusters could be merged (\eg{}, ``add water'' and ``use water'') or some actions should not belong in some of the clusters (\eg{}, ``put engine oil'' and ``paint with oil'').
Third, actions can be too ambiguous (``use water'') or too broad (\eg{}, ``add ingredient'').

\begin{figure*}
    \includegraphics{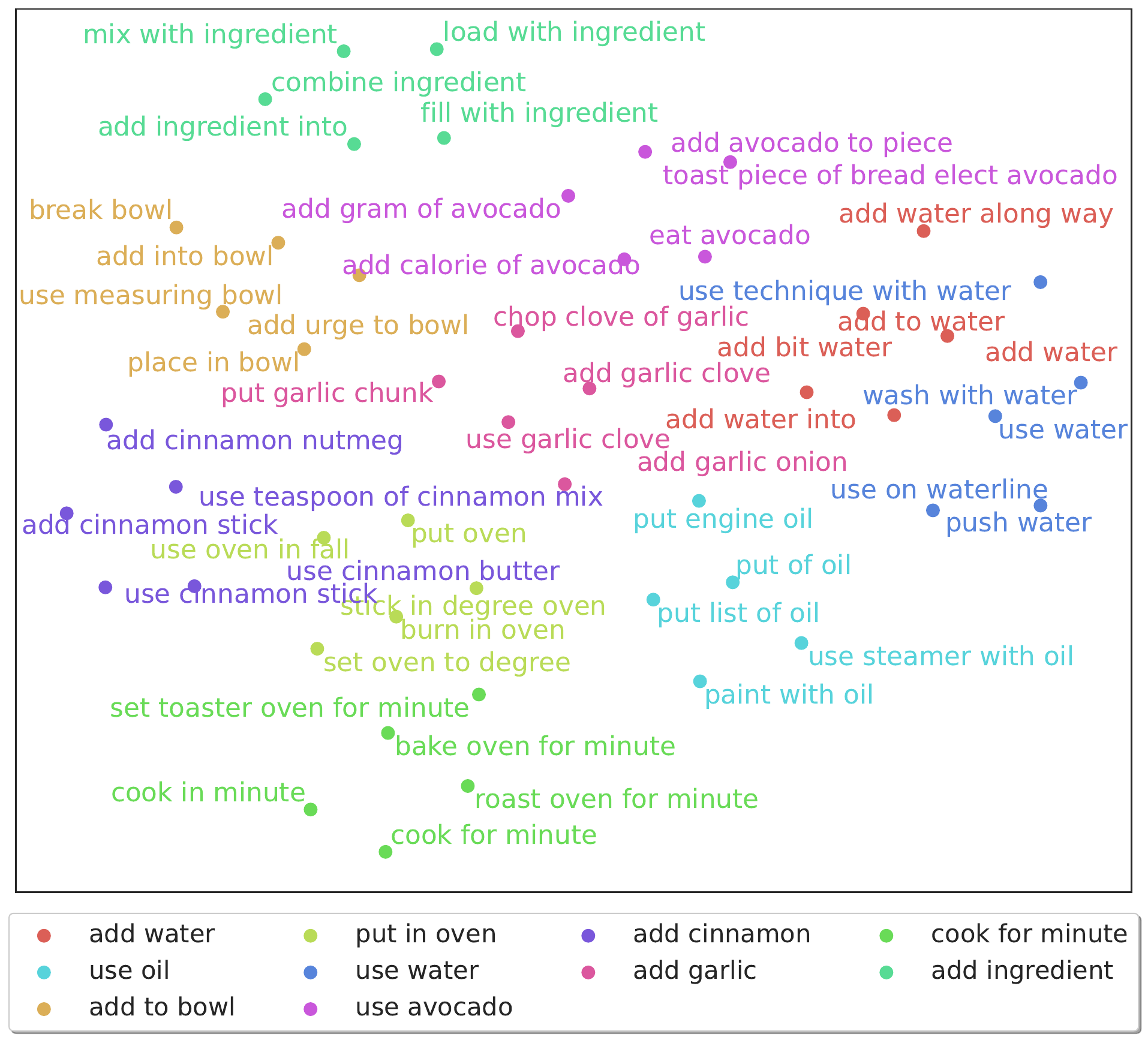}
    \caption{The t-SNE representation~\cite{Maaten08visualizingdata} of the ten most frequent action clusters in our dataset. Each color represents a different action cluster. Best viewed in color.}
    \label{fig:cluster}
\end{figure*}

\end{document}